\newif\ifisTR

\isTRtrue

\documentclass{article} 

\usepackage{fullpage}

\usepackage{microtype}
\usepackage{graphicx}
\usepackage{booktabs} 
\usepackage{placeins}
\usepackage{algorithm}
\usepackage{algorithmic}
\usepackage{subcaption}
\usepackage{tcolorbox}
\usepackage{nicefrac}  
\usepackage{enumitem}
\usepackage{multirow}
\usepackage{colortbl}
\usepackage{xspace}
\usepackage{wrapfig}
\usepackage{amsmath}
\usepackage{amssymb}
\usepackage{mathtools}
\usepackage[export]{adjustbox}
\usepackage{amsthm}
\usepackage{xcolor}

\definecolor{CColor}{rgb}{0.01,0.31,0.59}
\definecolor{GGray}{rgb}{0.80,0.90,1}
\definecolor{Shady}{rgb}{0.9,0.9,0.9}
\definecolor{kaistblue}{RGB}{20,135,200}
\definecolor{kaistdarkblue}{RGB}{0,65,145}
\definecolor{urbanablue}{RGB}{19,41,75}
\definecolor{urbanaorange}{RGB}{232,74,39}
\definecolor{drp}{rgb}{0.53,0.15,0.34}
\usepackage[plainpages=false,pdfpagelabels,colorlinks=true,linkcolor=CColor,citecolor=CColor,urlcolor=CColor]{hyperref}

\usepackage{tikz}

\usepackage[capitalize,noabbrev]{cleveref}

\usepackage{overpic}

\theoremstyle{plain}

\theoremstyle{definition}

\theoremstyle{remark}

\definecolor{mygray}{gray}{0.85}
\definecolor{LightBlue}{cmyk}{0.06, 0.03, 0.01, 0.0}

\usepackage[textsize=tiny]{todonotes}

\usepackage[round,semicolon,sort]{natbib}

\newcommand{\ourmethod}{AutoSurvey2\xspace}

\renewcommand{\cite}[1]{\citep{#1}}
\title{AutoSurvey2: Empowering Researchers with Next Level Automated Literature Surveys}
\date{}
\author{
Siyi Wu$^*$\textsuperscript{1},
XinLiang Chia$^*$\textsuperscript{2},
Ziqian Bi$^*$\textsuperscript{2,3},
Leyi Zhao\textsuperscript{3},\\
Tianyang Wang\textsuperscript{4},
Junhao Song\textsuperscript{5},
Yichao Zhang\textsuperscript{6},
Keyu Chen\textsuperscript{6},
Benji Peng\textsuperscript{7},
Xinyuan Song\textsuperscript{†8}\\[4pt]
\textsuperscript{1}University of Texas at Arlington, USA \\
\textsuperscript{2}AI Agent Lab, USA \\
\textsuperscript{3}Indiana University, USA \\
\textsuperscript{4}Ohio State University, USA \\
\textsuperscript{5}Imperial College London, UK \\
\textsuperscript{6}Georgia Institute of Technology, USA \\
\textsuperscript{7}Appcubic, USA \\
\textsuperscript{8}Emory University, USA \\
}

\begin{document}
\maketitle

\def\thefootnote{*}\footnotetext{Equal contribution.}
\def\thefootnote{†}\footnotetext{Correspondence to: Xinyuan Song ,xsong30@emory.edu}

\renewcommand{\thefootnote}{\arabic{footnote}}

%

\begin{abstract}
The rapid growth of research literature, particularly in large language models (LLMs), has made producing comprehensive and current survey papers increasingly difficult. This paper introduces \ourmethod, a multi-stage pipeline that automates survey generation through retrieval-augmented synthesis and structured evaluation. The system integrates parallel section generation, iterative refinement, and real-time retrieval of recent publications to ensure both topical completeness and factual accuracy. Quality is assessed using a multi-LLM evaluation framework that measures coverage, structure, and relevance in alignment with expert review standards. Experimental results demonstrate that \ourmethod consistently outperforms existing retrieval-based and automated baselines, achieving higher scores in structural coherence and topical relevance while maintaining strong citation fidelity. By combining retrieval, reasoning, and automated evaluation into a unified framework, \ourmethod provides a scalable and reproducible solution for generating long-form academic surveys and contributes a solid foundation for future research on automated scholarly writing. All code and resources are available at \url{https://github.com/annihi1ation/auto_research}.
\end{abstract}


\maketitle

\section{Introduction}
\label{sec:intro}

Survey papers play a central role in consolidating and disseminating knowledge across rapidly evolving research domains, offering comprehensive overviews of key advances, emerging trends, and open challenges~\cite{pouyanfar2018survey,chang2023survey,zhao2023survey,khan2022transformers}. In fields such as artificial intelligence and large language models (LLMs)~\cite{lecun2015deep,goodfellow2016deep,achiam2023gpt,kirillov2023segment}, the exponential increase in publication volume has made the preparation of high-quality surveys increasingly demanding. Although recent progress in LLMs~\cite{achiam2023gpt,touvron2023llama} enables long-context text generation~\cite{chen2023extending,chen2023longlora,wang2024augmenting}, directly employing them for survey writing remains challenging for three major reasons.

First, context window limitations~\cite{liu2024lost,kaddour2023challenges,shi2023large,li2023long,li2024long} restrict LLMs from processing large-scale literature corpora. Comprehensive surveys typically integrate hundreds of papers, requiring tens of thousands of tokens—far beyond the stable generation capacity of current models.  
Second, depending solely on an LLM’s internal knowledge is insufficient for constructing complete and up-to-date reviews~\cite{wang2023surveyfact,ji2023survey,shao2024assisting}, particularly for newly published works that may be absent from training data. This increases the risk of outdated or fabricated citations.  
Third, the lack of standardized evaluation protocols for LLM-generated academic content~\cite{pandalm2024,zheng2024judging,yu2024kieval} makes quality assurance difficult, as expert evaluation is costly, subjective, and hard to scale.

To address these challenges, we propose \ourmethod, an end-to-end automated framework for generating academic survey papers. The system integrates multi-stage retrieval, modular content generation, and automated evaluation into a cohesive pipeline (Section~\ref{sec:method}). It operates in four coordinated stages: (1) survey outline construction through retrieval-guided planning, (2) section-level literature retrieval and analysis using high-recall embeddings, (3) section content generation grounded in retrieved evidence, and (4) final post-processing and assembly into a publication-ready \LaTeX{} document. The pipeline is fully modular and compatible with multiple LLM providers, enabling scalable and reproducible survey generation.

Beyond generation, \ourmethod incorporates two key mechanisms that directly address prior limitations: a real-time retrieval-augmented generation strategy~\cite{gao2023retrieval,lewis2020retrieval,jiang2023active} that ensures integration of recent literature, and a multi-LLM evaluation framework~\cite{zheng2024judging,pandalm2024,yu2024kieval} that automatically assesses coverage, structure, and relevance in alignment with expert review criteria. Together, these components enable the system to produce surveys that are both comprehensive and stylistically coherent.

Empirical studies demonstrate that \ourmethod achieves consistent improvements in content organization, topic relevance, and factual coverage compared with existing automated baselines. The generated surveys exhibit strong structural coherence and accurate citation grounding, approaching the quality of human-written reviews while remaining reproducible and scalable.

The main contributions of this work are as follows:
\begin{itemize}
    \item We present \ourmethod, a modular and fully automated pipeline for generating long-form academic surveys through retrieval, structured planning, and large-model synthesis.
    \item We design a graph-based execution framework that decomposes survey writing into four coordinated stages, improving scalability and consistency.
    \item We demonstrate through experiments that the proposed system produces coherent, well-organized, and factually grounded survey papers, providing a reproducible foundation for automated academic writing.
\end{itemize}

In summary, this work contributes a unified framework that combines retrieval, reasoning, and evaluation for scalable and reliable survey generation, bridging the gap between automated synthesis and expert-level academic authorship.

\section{Related Work}

\textbf{Auto literature survey generation via LLMs.} Recent work has begun harnessing large language models (LLMs) to automate the writing of literature surveys. AutoSurvey~\cite{auto1} introduced a structured pipeline for LLM-driven survey generation to address the exploding volume of publications in rapidly evolving fields. The first version of AutoSurvey employs multiple specialised LLMs in a staged framework: relevant papers are initially retrieved via embedding-based search, and an outline is produced with an outline-generation LLM. Each section and subsection is then generated by separate, topic-specific LLM modules. The approach explicitly tackles the challenges of (1) limited context windows (through retrieval and chunked generation) and (2) incomplete parametric knowledge (via real-time Retrieval-Augmented Generation, RAG)~\citep{auto2, auto3, auto4}. AutoSurvey further integrates dedicated revision modules to enhance section coherence and introduces an evaluation mechanism in which multiple LLMs act as judges to score generated content for citation quality, coverage, and logical structure~\citep{auto5}. This modular pipeline enables near-comprehensive coverage and rigorous quality control, as demonstrated by extensive benchmarking against human-written surveys~\cite{auto6, auto7}.

In parallel, the Auto-survey Challenge established a shared task to systematically evaluate LLMs' capabilities to both compose and peer-review survey articles~\cite{auto8}. In this challenge, LLM-based systems are required to autonomously write survey drafts and participate in an iterative, LLM-mediated peer-review process. Systems are evaluated on writing quality, factual accuracy, citation relevance, and review insight, with human organisers providing ground-truth assessments. These works collectively position LLMs as collaborative tools for survey generation, able to accelerate drafting while still requiring human validation to ensure scientific rigour.

\textbf{LLMs for data annotation and scientific workflows.} Beyond survey writing, LLMs have increasingly been integrated into broader scientific research workflows, including data annotation, survey design, and automated evaluation. Recent surveys~\cite{sc1, sc2, sc3} outline how LLMs are applied across the research lifecycle, which from hypothesis generation and experimental planning to manuscript drafting and peer review. For data annotation, LLMs such as GPT-4.1 and Llama 4 are leveraged to generate labels or synthetic data at scale, reducing manual effort but raising challenges in annotation consistency and quality assurance~\cite{sc4, sc5, sc6}. Quality control typically requires hybrid human-in-the-loop approaches, where human experts validate or curate LLM-generated outputs. In survey design, LLMs can automatically draft questionnaire items and suggest answer choices ~\cite{sc7}. However, expert intervention remains essential for methodological soundness and contextual adaptation.

A further line of research investigates LLMs as evaluators or judges in scientific workflows. Also some research assess the use of LLMs for summarising and evaluating open-ended survey responses~\cite{sc8}, showing that LLM-based ratings align moderately with human judgments, but may miss subtle content or context-specific nuances. These studies demonstrate both the scalability of LLMs for automating tedious research tasks and their current limitations in factual verification, nuanced reasoning, and domain-specific reliability. The prevailing consensus is that LLMs are most effective as assistive tools in scientific research when complemented by human expertise for final validation and interpretation.
\section{Methodology}
\label{sec:method}

\subsection{Overview}
\label{sec:method_overview}

\begin{figure}[ht!]
    \centering
    \includegraphics[width=\linewidth]{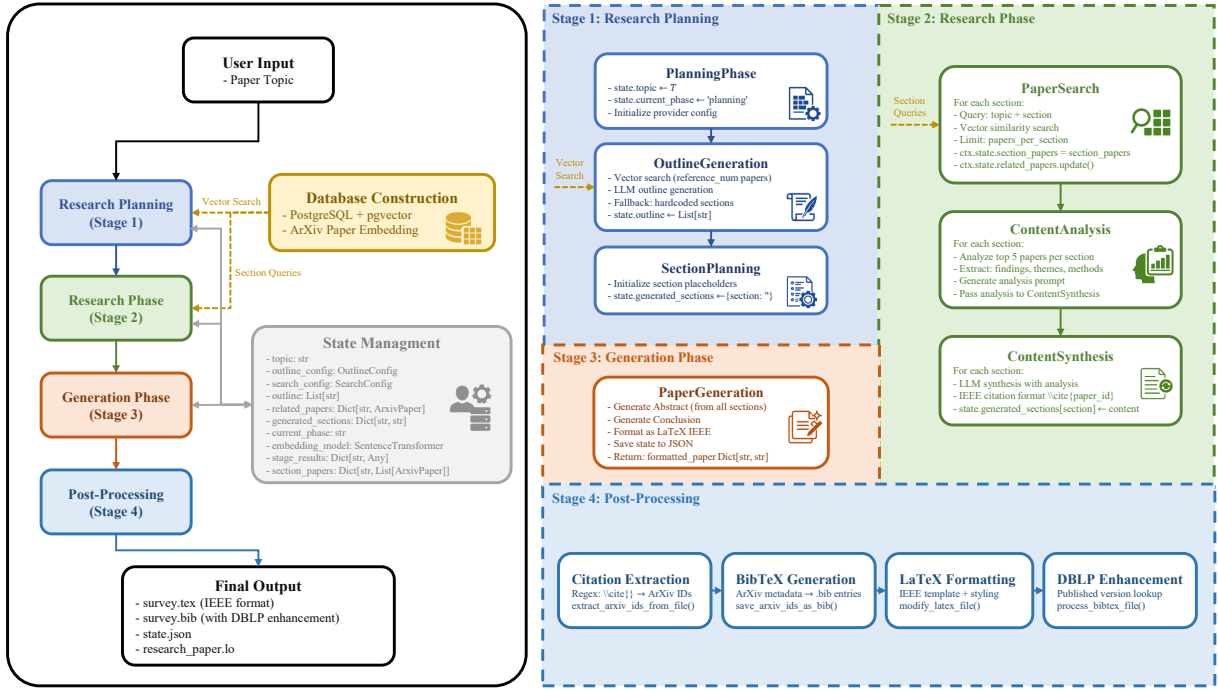}
    \caption{Overall workflow of \ourmethod. The system operates through four sequential stages organized as a graph-based state machine: \textit{Database Construction} (prerequisite), \textit{Research Planning} (Stage 1), \textit{Research Phase} (Stage 2), \textit{Generation Phase} (Stage 3), and \textit{Post-processing} (Stage 4). Each stage comprises specialized functional nodes that communicate through a shared global state $\Sigma$.}
    \label{fig:pipeline}
\end{figure}

Figure~\ref{fig:pipeline} illustrates the architecture of \ourmethod, an automated framework designed to generate comprehensive academic survey papers. Building upon recent advances in automated literature review generation \cite{cho2018traingcohernt, bosselut2018awardcoherent, wang2019paperrobot}, the system integrates large language models (LLMs), semantic vector retrieval, and a modular graph-based execution mechanism to emulate the scholarly writing process in a structured and reproducible manner. The workflow proceeds through four interconnected stages that together form a complete pipeline for survey generation. It begins with the construction of a semantically indexed database, which serves as the foundation for all subsequent operations. In this stage, metadata from ArXiv is collected and processed into dense embeddings using transformer-based encoders \cite{nussbaum2024nomic}. The resulting representations are stored in a PostgreSQL database equipped with \texttt{pgvector} support, enabling efficient similarity searches that underpin the retrieval capabilities of the entire system.

Once the database is established, the process transitions to the research planning stage, where a user-defined topic $\mathcal{T}$ serves as input. The system retrieves a relevant subset of papers and employs an LLM to analyze them, producing a structured outline $\mathcal{O} = \{s_1, s_2, \ldots, s_m\}$ that captures the main themes, methods, and research directions associated with $\mathcal{T}$ \cite{shao2024assisting}. Each element $s_i$ in the outline is further explored during the research phase, which retrieves section-specific papers, identifies their methodological contributions, and synthesizes this information into structured prompts. These prompts guide the subsequent generation phase, where the system produces coherent academic text with inline IEEE-style citations \cite{balepur2023expository}. The final post-processing stage refines the manuscript by extracting citation data, generating BibTeX entries, and applying IEEE formatting. During this process, DBLP is queried to replace preprint references with their formally published counterparts, thereby enhancing citation accuracy and overall document quality.

All stages are orchestrated through a graph-based state machine that ensures systematic and reproducible execution. The system employs a directed acyclic graph (DAG) in which each node corresponds to a specialized operation—such as \texttt{OutlineGeneration}, \texttt{PaperSearch}, or \texttt{ContentSynthesis}—and edges encode task dependencies. These nodes share a global state $\Sigma$ that stores intermediate data including the topic, outline, retrieved papers, generated sections, and configuration parameters. Maintaining this shared state guarantees consistency and traceability throughout the workflow while allowing modular development, testing, and debugging of individual components. The DAG-based design thus provides both flexibility and stability, enabling independent evolution of system modules without compromising the integrity of the end-to-end process. The following subsections detail the technical implementation of each stage, outlining the algorithms, models, and design principles that enable high-quality automated survey generation.

\subsection{Database Construction}
\label{sec:method_database}

Before survey generation, a semantically indexed repository of academic papers is constructed to serve as the retrieval foundation for all downstream stages. Similar large-scale retrieval infrastructures have been used in prior works on semantic vector search and text embedding retrieval \cite{shan2018recurrentbinary, fang2020beyondlexical, seyedmonir2024vectorsearch}. This repository transforms raw ArXiv metadata into a vector-searchable database optimized for large-scale semantic similarity queries. Metadata is retrieved via the Kaggle API using the official \texttt{Cornell-University/arxiv} dataset, which contains comprehensive bibliographic records including paper identifiers, titles, abstracts, author lists, categories, and version histories. The system authenticates using JSON-based credentials and downloads the complete dataset snapshot, which is then filtered to retain only papers in AI- and ML-related categories:
\begin{equation}
\mathcal{D}_{\text{filtered}} = \{ p \in \mathcal{D}_{\text{raw}} \mid \texttt{category}(p) \cap \{\text{cs.AI, cs.CL, cs.CV, cs.LG, stat.ML}\} \neq \emptyset \}.
\end{equation}
This filtering ensures that the resulting collection remains domain-specific while preserving comprehensive coverage across artificial intelligence and machine learning research areas, following common practices in embedding-based scientific indexing \cite{liu2024relevancefiltering}.

For each paper $p \in \mathcal{D}_{\text{filtered}}$, a dense semantic embedding is generated by encoding its abstract using a transformer-based sentence encoder. Specifically, we employ the \texttt{nomic-embed-text-v2-moe} model \cite{nussbaum2025trainingsparsemixtureexperts}, a mixture-of-experts architecture that produces 768-dimensional representations:
\begin{equation}
\mathbf{e}_p = f_{\theta}(\texttt{abstract}_p), \quad \mathbf{e}_p \in \mathbb{R}^{768},
\end{equation}
where $f_{\theta}$ denotes the pretrained encoder. This design follows the growing trend of mixture-of-expert embedding models for high-dimensional semantic representation \cite{wu2025sitemb, nussbaum2025theoreticallimitations}. The model runs on GPU (CUDA or MPS) when available and gracefully falls back to CPU otherwise. Embeddings are computed in batches of 10,000 papers to balance memory efficiency and throughput, a strategy widely adopted in scalable embedding pipelines \cite{shan2018recurrentbinary}.

These representations are stored in a PostgreSQL database extended with the \texttt{pgvector} module, which enables native indexing and querying of high-dimensional vectors. Each record includes standard metadata fields such as identifiers, titles, authors, abstracts, categories, and structured version histories, along with an \texttt{embedding} field that serializes 768-dimensional vectors into PostgreSQL array format. This setup supports efficient semantic search through the inner-product operator \texttt{<->}, providing the retrieval backbone for subsequent stages of the system \cite{seyedmonir2024vectorsearch}.

To facilitate incremental updates while minimizing redundant computation, the ingestion pipeline first checks for existing paper identifiers before insertion:
\begin{equation}
\mathcal{D}_{\text{new}} = \mathcal{D}_{\text{filtered}} \setminus \{ p \mid \texttt{id}_p \in \mathcal{D}_{\text{existing}} \},
\end{equation}
where $\mathcal{D}_{\text{existing}}$ represents the set of papers already present in the database. Only new entries $\mathcal{D}_{\text{new}}$ undergo embedding generation and insertion, which substantially reduces processing time during incremental updates. The ingestion process is continuously monitored via detailed logging and progress tracking, ensuring transparency and fault tolerance. The final repository $\mathcal{D}_{\text{arxiv}}$ comprises hundreds of thousands of papers with precomputed embeddings, enabling sub-second retrieval of semantically related documents and forming the essential retrieval layer that powers automated survey generation \cite{fang2020beyondlexical, nussbaum2025trainingsparsemixtureexperts}.

\subsection{Stage 1: Research Planning}
\label{sec:method_planning}

The research planning stage transforms a user-specified topic into a structured survey blueprint through three sequential nodes—\texttt{PlanningPhase}, \texttt{OutlineGeneration}, and \texttt{SectionPlanning}—executed within the graph-based state machine. Each node reads from and writes to the shared global state \(\Sigma\), maintaining consistency and traceability throughout the process. The workflow begins at the \texttt{PlanningPhase} node, which initializes the system state with the user-defined topic:
\begin{equation}
\Sigma.\texttt{topic} \leftarrow \mathcal{T}, \quad \Sigma.\texttt{current\_phase} \leftarrow \text{``planning''}.
\end{equation}
During this initialization, the system also loads configuration parameters that define the operating environment, including the LLM provider (e.g., Ollama, OpenAI, or OpenRouter), the model identifier, and retrieval hyper-parameters. These settings are stored in $\Sigma.\texttt{stage\_results}$ to ensure consistent downstream behavior and reproducibility across all stages of the pipeline.

Once initialization is complete, the \texttt{OutlineGeneration} node constructs a hierarchical outline \(\mathcal{O} = \{s_1, s_2, \ldots, s_m\}\) that determines the overall organizational structure of the survey. This step combines large-scale semantic retrieval with LLM-based theme identification, following recent methods in automated survey generation \cite{wang2024autosurvey, lai2024stepbystep}. A reference corpus \(\mathcal{D}_{\text{ref}}\) of \(K_{\text{ref}}\) papers (default \(K_{\text{ref}} = 1500\)) is retrieved by computing the cosine similarity between the topic embedding and the embeddings of all papers in the database:
\begin{equation}
\mathcal{D}_{\text{ref}} = \underset{D' \subset \mathcal{D}_{\text{arxiv}},   |D'| = K_{\text{ref}}}{\arg\max} \sum_{p \in D'} \frac{\mathbf{e}_{\mathcal{T}}^\top \mathbf{e}_p}{\|\mathbf{e}_{\mathcal{T}}\| \|\mathbf{e}_p\|}.
\end{equation}
The titles and abstracts of \(\mathcal{D}_{\text{ref}}\) are then provided as context to an LLM, which is prompted to identify major research themes, methodological paradigms, and open challenges related to the target topic. The prompt constrains the output to generate exactly \(m\) sections (default \(m = 8\)), omits subsections, and enforces a numbered list format \cite{lai2024stepbystep}. The resulting outline is parsed using regular expressions to extract section titles and stored as \(\Sigma.\texttt{outline}\), which serves as the structural foundation for subsequent stages.

Finally, the \texttt{SectionPlanning} node prepares the internal representation for each section defined in \(\mathcal{O}\) by initializing empty content placeholders:
\begin{equation}
\Sigma.\texttt{generated\_sections} \leftarrow \{ s_i \mapsto \text{``''} \mid s_i \in \mathcal{O} \}.
\end{equation}
This ensures that every section has a corresponding entry in the global state even before generation begins, preserving structural completeness. Each section is also annotated with metadata describing its order and hierarchical relationships, facilitating coordinated retrieval and generation in later stages. Upon completion, the global state \(\Sigma\) contains the finalized outline \(\mathcal{O}\), initialized section placeholders, and all configuration parameters required for content synthesis. This structured planning process establishes a coherent organizational backbone grounded in relevant literature, ensuring that the generated survey adheres to a logically consistent and thematically comprehensive design.

\subsection{Stage 2: Research Phase}
\label{sec:method_research}

The research phase retrieves, analyzes, and synthesizes relevant literature for each section defined in the outline \(\mathcal{O}\). It consists of three sequential nodes—\texttt{PaperSearch}, \texttt{ContentAnalysis}, and \texttt{ContentSynthesis}—that progressively transform raw paper collections into structured content suitable for LLM-based generation. For each section \(s_i \in \mathcal{O}\), the \texttt{PaperSearch} node performs targeted retrieval to identify papers most relevant to that section’s theme. The search query is formed by concatenating the main topic with the section title:
\begin{equation}
q_i = \mathcal{T} \oplus s_i,
\end{equation}
where \(\oplus\) denotes string concatenation. The query is embedded using the same encoder \(f_{\theta}\) employed during database construction, yielding a vector representation \(\mathbf{e}_{q_i} \in \mathbb{R}^{768}\). The system then retrieves the top-\(K_{\text{sec}}\) most similar papers (default \(K_{\text{sec}} = 20\)) from the database:
\begin{equation}
\mathcal{P}_i = \underset{P' \subset \mathcal{D}_{\text{arxiv}},   |P'| = K_{\text{sec}}}{\arg\max} \sum_{p \in P'} \text{sim}(\mathbf{e}_{q_i}, \mathbf{e}_p),
\end{equation}
where \(\text{sim}(\cdot, \cdot)\) represents cosine similarity computed via PostgreSQL’s \texttt{<->} operator. The retrieved papers \(\mathcal{P}_i\) are stored in \(\Sigma.\texttt{section\_papers}[s_i]\), and their identifiers are aggregated in \(\Sigma.\texttt{related\_papers}\) to maintain a global record of all referenced sources across sections.

After retrieval, the \texttt{ContentAnalysis} node examines the top-\(k\) papers (default \(k = 5\)) from each set \(\mathcal{P}_i\) to extract structured insights. To balance computational efficiency with informational richness, only the most relevant subset of papers undergo detailed analysis. The node constructs a context string by concatenating the titles and abstracts of selected papers:
\begin{equation}
C_i = \bigoplus_{p \in \mathcal{P}_i[:k]} (\texttt{title}_p \oplus \texttt{abstract}_p).
\end{equation}
This context is passed to an LLM along with a structured prompt that requests identification of key contributions, common methodologies, contrasting perspectives, technical nuances, and existing limitations. The model’s response, denoted as \(A_i\), provides a concise synthesis of these dimensions and is stored in \(\Sigma.\texttt{section\_analysis}[s_i]\), forming a distilled knowledge base for the subsequent synthesis stage.

Building on these analyses, the \texttt{ContentSynthesis} node generates draft text for each section by combining the analytical summary \(A_i\), the retrieved paper set \(\mathcal{P}_i\), and explicit stylistic and structural constraints. The LLM is instructed to write in a formal academic tone, ensure logical flow, use IEEE-style inline citations via \verb|\cite{paper_id}| commands, and maintain cohesive coverage without introducing subsections. The generated section content is formalized as:
\begin{equation}
c_i = \text{LLM}(\mathcal{T}, s_i, \mathcal{P}_i, A_i \mid \theta, \rho_{\text{synthesis}}),
\end{equation}
where \(\theta\) represents the model parameters and \(\rho_{\text{synthesis}}\) denotes the synthesis prompt template. The resulting drafts are stored in \(\Sigma.\texttt{generated\_sections}[s_i]\), while sections such as “Abstract” or “Conclusion” are deferred for later generation. This staged process ensures that each section is grounded in relevant literature, enriched with contextual understanding, and stylistically consistent across the entire survey \cite{wei2025plangenllms}.

\subsection{Stage 3: Generation Phase}
\label{sec:method_generation}

The generation phase, implemented by the \texttt{PaperGeneration} node, produces the final abstract and conclusion sections, formats all content into IEEE‐compliant \LaTeX, and saves the complete system state for reproducibility. After all main sections \(\{c_i\}_{i=1}^{m}\) have been generated, the system synthesizes a concise abstract \(\mathcal{A}\) that summarizes the survey’s motivation, scope, and principal contributions. To capture the overall essence of the work while adhering to token constraints, a condensed summary is constructed by concatenating section titles with truncated content segments:
\begin{equation}
\mathcal{S}_{\text{summary}} = \bigoplus_{i=1}^{m} (s_i \oplus c_i[:500]),
\end{equation}
where \(c_i[:500]\) denotes the first 500 characters of section \(c_i\). This summary is provided as context for an LLM prompt that generates a 250–300 word abstract following the conventional academic structure of motivation, methods, findings, and contributions \cite{instruct_llm_survey_2024}. The resulting abstract is defined as
\begin{equation}
\mathcal{A} = \text{LLM}(\mathcal{T}, \mathcal{S}_{\text{summary}} \mid \rho_{\text{abstract}}),
\end{equation}
where \(\rho_{\text{abstract}}\) specifies the abstract generation prompt, and the output is stored in \\ \(\Sigma.\texttt{generated\_sections}[\text{"Abstract"}]\). Following the same approach, the system generates a conclusion \(\mathcal{C}\) by prompting the LLM to compose closing remarks that summarise key insights, highlight the current research landscape, discuss future directions, and provide a reflective synthesis of the field. The process is formalised as
\begin{equation}
\mathcal{C} = \text{LLM}(\mathcal{T}, \mathcal{S}_{\text{summary}} \mid \rho_{\text{conclusion}}),
\end{equation}
with the output constrained to 400–500 words and stored in \(\Sigma.\texttt{generated\_sections}[\text{"Conclusion"}]\), completing the full set of survey sections.

Once textual generation is finalised, all sections are transformed into IEEE‐compliant \LaTeX\ through the \texttt{\_format\_latex} method. This procedure constructs the full paper structure, including the preamble, title block, abstract environment, section organisation, and citation normalisation. The preamble defines the document class and required packages (e.g., \texttt{cite}, \texttt{amsmath}, \texttt{graphicx}), while the title block includes placeholders for author affiliations and funding acknowledgments. The abstract is wrapped in the standard \verb|\begin{abstract}...\end{abstract}| environment with appropriate keywords, and each section \(s_i\) and its content \(c_i\) are rendered as \verb|\section{$s_i$}| blocks with special characters escaped for compatibility. Inline citations are standardised to the \verb|\cite{...}| format, replacing any temporary placeholders. The formatted output is represented as a dictionary mapping section names to their corresponding \LaTeX\ strings:
\begin{equation}
\mathcal{F}_{\text{paper}} = \{ \text{"Preamble"} \mapsto \ldots,  \text{"Abstract"} \mapsto \ldots,  s_i \mapsto \ldots \}.
\end{equation}

To ensure reproducibility and facilitate downstream analysis, the complete system state is serialized to JSON and saved to the \texttt{result/} directory. The saved state includes metadata, topic definition, outline, related papers, retrieved paper sets, generated sections, and the formatted paper:
\begin{equation}
\mathcal{J}_{\text{state}} = \{\texttt{metadata}, \texttt{topic}, \texttt{outline}, \texttt{related\_papers}, \texttt{section\_papers}, \texttt{generated\_sections}, \texttt{final\_paper}\}.
\end{equation}
Within this snapshot, \texttt{metadata} records timestamps, model identifiers, and configuration parameters, enabling precise reconstruction of experimental conditions. This serialized state allows individual sections to be modified or regenerated independently without re‐running the full pipeline. The \texttt{PaperGeneration} node ultimately outputs \(\mathcal{F}_{\text{paper}}\), which is passed to the post-processing stage for citation refinement and final compilation \cite{automated_review_generation_2025}.

\subsection{Stage 4: Post-processing}
\label{sec:method_postprocessing}

The post-processing stage refines the generated survey into a publication-ready document through a series of automated transformations, including citation extraction, BibTeX generation, LaTeX formatting enhancement, and citation quality improvement via DBLP integration. Implemented by the \texttt{post\_process\_paper} function, this stage operates directly on the generated \texttt{.tex} file to ensure that all citations, formatting elements, and metadata conform to publication standards. The process begins with citation extraction, where the system parses the LaTeX source to identify all citation keys referenced in the text. Using regular expression matching, every occurrence of the \verb|\cite{}| command is extracted:
\begin{equation}
\mathcal{C}_{\text{keys}}
= \{  c \mid c \in \operatorname{match}(\text{\texttt{\string\cite\{\}}}),  c \neq \varnothing  \}.
\end{equation}

yielding a complete set of unique citation identifiers (typically ArXiv IDs). Multiple citations within a single command are split on commas and whitespace to ensure full coverage. For each key \(c \in \mathcal{C}_{\text{keys}}\), the system retrieves the corresponding metadata from the ArXiv snapshot and converts it into an IEEE‐style BibTeX entry following the \texttt{@misc} schema:
\begin{equation}
\texttt{BibEntry}(c) = g_{\phi}(\texttt{metadata}_c),
\end{equation}
where \(g_{\phi}\) maps metadata fields such as authors, title, year, and categories into structured BibTeX format. Author names are formatted in “FirstName LastName” order using the parsed author field, and publication years are inferred from ArXiv ID prefixes. The resulting entries are written to a \texttt{.bib} file co-located with the LaTeX source, ensuring full bibliographic consistency.

Once the citation data have been assembled, the system refines the generated \LaTeX\ source to ensure IEEE-compliant formatting. The \texttt{modify\_latex\_file} function performs a series of structural enhancements: it replaces the default \texttt{cite} package with \texttt{natbib} to enable advanced citation control, and integrates additional packages such as \texttt{tikz}, \texttt{forest}, \texttt{booktabs}, and \texttt{hyperref} to support high-quality typesetting. Hyperlink colours are configured in dark blue to improve readability and maintain a professional appearance. The title block is automatically derived from the filename, converted to title case while preserving acronyms (e.g., “MLLM”), and suffixed with “A Comprehensive Survey.” The abstract is enclosed within the standard \verb|\begin{abstract}...\end{abstract}| environment, and each section \(s_i\) is formatted as a \verb|\section{$s_i$}| block with escaped special characters for syntactic safety. The bibliography is appended through \verb|\bibliographystyle{apalike}| and \verb|\bibliography{filename}| commands, while consistent vertical spacing is inserted before section headers to enhance visual clarity. Collectively, these adjustments ensure that the final \LaTeX\ document compiles seamlessly and conforms to professional academic presentation standards \cite{outline_guided_generation_2025}.

To further enhance citation quality, the system integrates with the DBLP database to identify formally published versions of ArXiv pre-prints. For each BibTeX entry, the \texttt{process\_bibtex\_file} function constructs a DBLP query from the title and first author, retrieves the search results, and parses available BibTeX entries. If a formally published version is found, the system replaces the ArXiv entry while retaining the original cite key for consistency; otherwise, the original entry is preserved with a comment indicating that only a pre-print is available. This conditional replacement can be formalised as
\begin{equation}
\texttt{BibEntry}_{\text{final}}(c) = 
\begin{cases}
\texttt{BibEntry}_{\text{DBLP}}(c) & \text{if published version exists},\\
\texttt{BibEntry}_{\text{ArXiv}}(c) & \text{otherwise.}
\end{cases}
\end{equation}
Duplicate keys are subsequently removed to maintain a clean bibliography. Finally, the post-processing pipeline assembles all outputs into a finalised directory:
\begin{equation}
\mathcal{D}_{\text{output}} = \{\texttt{survey.tex}, \texttt{survey.bib}, \texttt{figs/}\},
\end{equation}
ready for compilation using \texttt{pdflatex} and \texttt{bibtex}. The resulting PDF constitutes a fully formatted academic survey with verified references and publication-ready IEEE styling.
\section{Prompt Design}
\label{sec:prompt_design}

Effective prompt engineering is essential for ensuring that large language models (LLMs) generate coherent, factual, and academically rigorous content throughout the \ourmethod pipeline. Each stage of the workflow is guided by a compact and structured prompt that defines the task, provides contextual grounding, and enforces explicit stylistic and structural constraints. During outline generation, the model acts as a survey‐outline constructor that receives the topic and a reference corpus containing paper titles and abstracts. It is instructed to produce a numbered list of section titles that follows standard academic structure while maintaining logical progression, balanced coverage, and stylistic consistency. For each section, a subsequent analysis prompt examines the most relevant retrieved papers, summarizing key contributions, shared patterns, technical approaches, and challenges into a concise, structured synthesis that forms the basis for generation. The following synthesis prompt transforms these insights into coherent academic prose, ensuring formal tone, logical organization, and accurate IEEE‐style citations. Through these progressive stages, the system maintains factual grounding while producing text that integrates seamlessly within a LaTeX‐based academic workflow \cite{amatriain2024promptdesign, sahoo2024systematicsurvey}.

After all main sections are generated, meta‐section prompts are used to create the abstract and conclusion. The abstract prompt produces a 250–300 word summary highlighting the survey’s motivation, scope, and contributions, while the conclusion prompt generates a 400–500 word synthesis discussing main findings, research trends, and open challenges. Across all stages, the prompt design adheres to several shared principles: clear role specification to align the model’s behaviour with academic conventions; structured requirements that reduce ambiguity and ensure systematic coverage; contextual grounding in retrieved literature to prevent hallucinations; explicit formatting constraints that support deterministic parsing; and reasoning encouragement through step‐by‐step instructions to improve coherence. Collectively, these design strategies ensure that the generated survey maintains scholarly rigor, stylistic consistency, and factual reliability across all components of the automated writing pipeline \cite{chen2023unleashingprompts, li2023practicalzeroshot}.
\section{Experiments}
\label{sec:experiments}

The experiments were conducted to assess both the effectiveness and completeness of the proposed pipeline in generating long-form academic survey papers. We designed a reproducible and transparent setup to ensure that all evaluations could be reliably replicated and interpreted.

The system infrastructure employed PostgreSQL as the primary database \\ (\texttt{localhost:5451}, \texttt{research\_db}) to manage paper metadata and retrieval results, with all vector operations executed using the \texttt{nomic-ai/nomic-embed-text-v1.5} embedding model \cite{nussbaum2025trainingsparsemixtureexperts}. This model generates 768-dimensional representations that provide a strong balance between semantic accuracy and computational efficiency. GPT-4.1 served as the backbone language model for all generation tasks, including outline construction, section-level writing, and post-generation refinement. For retrieval-augmented generation, vector search was configured with a top-\(k\) of 100 and a similarity threshold of 0.7 to filter semantically relevant documents. Each survey generation process considered up to 1,500 reference papers, producing eight sections per survey, with 20 candidate papers retrieved per section to ensure sufficient topical coverage while maintaining computational tractability.

To evaluate the quality of generated surveys, we adopted an automated assessment framework based on a Judge Agent that formalizes expert review criteria into quantifiable metrics. The framework evaluates three fundamental dimensions: \textit{Coverage}, which measures the comprehensiveness of topic inclusion; \textit{Structure}, which assesses logical organization and narrative coherence; and \textit{Relevance}, which evaluates topical alignment and citation accuracy. Each dimension follows a 5-point Likert scale ranging from 1 (poor) to 5 (excellent), with detailed quality descriptors for intermediate levels. Evaluation prompts combine the full survey text, explicit scoring rubrics, and detailed descriptions of each criterion. The Judge Agent performs deterministic inference at zero temperature to ensure reproducibility, extracting structured numerical scores from the model output through pattern-based parsing. This automated evaluation strategy aligns with recent frameworks for LLM-based academic assessment and review automation \cite{automated_review_generation_2025, surveygen_quality_2025}.

Each experiment used ten representative topics across domains such as computer science, mathematics, and physics. All generated surveys were evaluated both individually and in comparison to baseline systems to assess content coverage, structural quality, and relevance of citations, consistent with evaluation methodologies proposed in large-scale survey generation benchmarks \cite{yan2025surveyforge, lai2024stepbystep}. The resulting scores were averaged across assessors and topics, providing a balanced view of overall system performance and enabling direct comparison across experimental configurations.
\section{Results}
\label{sec:results}

Table~\ref{table:comparison} presents a comparative evaluation between \ourmethod, AutoSurvey, and a Naive RAG baseline across three dimensions: \textit{Coverage}, \textit{Structure}, and \textit{Relevance}. Overall, \ourmethod achieves the highest average score of 4.76, outperforming all baselines in every category. Compared with AutoSurvey, the improvements in both \textit{Structure} and \textit{Coverage} indicate that the integration of graph-based planning and semantic retrieval enhances the coherence and comprehensiveness of the generated surveys. The Naive RAG approach, while capable of producing generally relevant content, exhibits weak structural consistency due to its lack of hierarchical topic decomposition and state tracking. These findings confirm that structured orchestration of LLM modules plays a decisive role in generating logically organized and contextually grounded survey papers.

\begin{table}[!ht]
\centering
\caption{Comparison between \ourmethod, AutoSurvey, and Naive RAG.}
\label{table:comparison}
\begin{adjustbox}{width=0.55\textwidth}
\begin{tabular}{c|cccc}
\toprule
\hline
Methods & Coverage & Structure & Relevance & Avg. \\ 
\midrule

Naive RAG & $4.46$ & $3.66$ & $4.73$ & $4.23$ \\

AutoSurvey & $4.66$ & $4.43$ & $4.86$ & $4.60$ \\

AutoSurvey2 (Ours) & $4.72$ & $4.68$ & $4.88$ & $4.76$ \\
\hline
\bottomrule
\end{tabular}
\end{adjustbox}
\end{table}

To further examine the contribution of key modules, we conducted an ablation study summarized in Table~\ref{table:ablation}. When the planning component was removed, the system experienced the most significant performance drop, particularly in \textit{Structure} (from 4.68 to 3.78), reflecting the importance of global topic decomposition for maintaining logical continuity across sections. Similarly, removing the refactoring stage led to decreased \textit{Structure} and \textit{Relevance}, as the absence of iterative refinement reduced inter-section alignment and stylistic uniformity. Despite these degradations, both ablated variants still outperform earlier baselines, demonstrating that the underlying retrieval and synthesis mechanisms remain robust. Together, these results highlight that the modular design of \ourmethod—especially its planning and refactoring components—is essential for achieving high-quality, academically coherent survey generation.

\begin{table}[!ht]
\centering
\caption{Ablation study results for \ourmethod with different components removed.}
\label{table:ablation}
\begin{adjustbox}{width=0.55\textwidth}
\begin{tabular}{c|cccc}
\toprule
\hline
Methods & Coverage & Structure & Relevance & Avg. \\ 
\midrule

AutoSurvey2 & $4.72$ & $4.68$ & $4.88$ & $4.76$ \\

AutoSurvey2 w/o planner & $4.51$ & $3.78$ & $4.77$ & $4.35$ \\

AutoSurvey2 w/o refactor & $4.69$ & $4.22$ & $4.79$ & $4.57$ \\
\hline
\bottomrule
\end{tabular}
\end{adjustbox}
\end{table}

\section{Conclusion}
\label{sec:conclusion}

This work presented \ourmethod, a multi-stage framework for generating long-form academic survey papers through structured planning, retrieval augmentation, and iterative refinement. By decomposing the writing process into sequential yet interdependent phases, the system effectively mitigates the limitations of large language models related to context length and incomplete domain knowledge. The integration of a retrieval-augmented database ensures timely inclusion of newly published research, thereby reducing citation obsolescence. Experimental results demonstrate that \ourmethod consistently outperforms prior baselines in terms of coverage, structural coherence, and relevance, achieving survey quality that approaches human-authored standards. These findings validate the effectiveness of combining semantic retrieval, modular planning, and LLM-driven synthesis for scalable academic writing.
\section{Limitations}
\label{sec:limitations}

Despite its strong performance, \ourmethod still faces several constraints. The system’s output quality is inherently dependent on the completeness and accuracy of the underlying literature database—missing or misclassified entries may cause important works to be overlooked. Moreover, although parallelized section generation accelerates the process, it can occasionally introduce stylistic inconsistencies that require additional refinement during post-processing. Finally, since the framework relies on large language models for both generation and evaluation, it inherits their known limitations, including potential factual inaccuracies, citation errors, or subtle biases derived from pretraining data. Addressing these issues will require tighter integration of verification modules and human-in-the-loop review mechanisms in future iterations.

\section{Ethical Statement}
\label{sec:ethics}

The automation of literature synthesis through large language models introduces both opportunities and ethical challenges. On one hand, such systems can democratize access to scholarly knowledge by enabling efficient exploration of vast research corpora. On the other hand, they may inadvertently propagate factual errors, biased interpretations, or incorrect citations if not carefully verified. To mitigate these risks, \ourmethod incorporates citation validation routines and encourages human review of all generated outputs prior to dissemination. Furthermore, we emphasize respect for intellectual property and data provenance when using public datasets. Ultimately, \ourmethod is designed not as a replacement for human scholarship, but as a tool to augment academic productivity and support critical, expert-driven research synthesis.

\bibliographystyle{ACM-Reference-Format}
\bibliography{bibfile}


\begin{thebibliography}{66}


\ifx \showCODEN    \undefined \def \showCODEN     #1{\unskip}     \fi
\ifx \showISBNx    \undefined \def \showISBNx     #1{\unskip}     \fi
\ifx \showISBNxiii \undefined \def \showISBNxiii  #1{\unskip}     \fi
\ifx \showISSN     \undefined \def \showISSN      #1{\unskip}     \fi
\ifx \showLCCN     \undefined \def \showLCCN      #1{\unskip}     \fi
\ifx \shownote     \undefined \def \shownote      #1{#1}          \fi
\ifx \showarticletitle \undefined \def \showarticletitle #1{#1}   \fi
\ifx \showURL      \undefined \def \showURL       {\relax}        \fi
\providecommand\bibfield[2]{#2}
\providecommand\bibinfo[2]{#2}
\providecommand\natexlab[1]{#1}
\providecommand\showeprint[2][]{arXiv:#2}

\bibitem[Achiam et~al\mbox{.}(2023)]%
        {achiam2023gpt}
\bibfield{author}{\bibinfo{person}{Josh Achiam}, \bibinfo{person}{Steven
  Adler}, \bibinfo{person}{Sandhini Agarwal}, \bibinfo{person}{Lama Ahmad},
  \bibinfo{person}{Ilge Akkaya}, \bibinfo{person}{Florencia~Leoni Aleman},
  \bibinfo{person}{Diogo Almeida}, \bibinfo{person}{Janko Altenschmidt},
  \bibinfo{person}{Sam Altman}, \bibinfo{person}{Shyamal Anadkat},
  {et~al\mbox{.}}} \bibinfo{year}{2023}\natexlab{}.
\newblock \bibinfo{title}{Gpt-4 technical report}.
\newblock


\bibitem[Amatriain(2024)]%
        {amatriain2024promptdesign}
\bibfield{author}{\bibinfo{person}{Xavier Amatriain}.}
  \bibinfo{year}{2024}\natexlab{}.
\newblock \bibinfo{title}{Prompt Design and Engineering: Introduction and
  Advanced Methods}.
\newblock
\showeprint[arxiv]{2401.14423}~[cs.CL]
\urldef\tempurl%
\url{https://arxiv.org/abs/2401.14423}
\showURL{%
\tempurl}


\bibitem[Balepur et~al\mbox{.}(2023)]%
        {balepur2023expository}
\bibfield{author}{\bibinfo{person}{Nishant Balepur}, \bibinfo{person}{Jie
  Huang}, {and} \bibinfo{person}{Kevin Chen-Chuan Chang}.}
  \bibinfo{year}{2023}\natexlab{}.
\newblock \bibinfo{title}{Expository text generation: Imitate, retrieve,
  paraphrase}.
\newblock


\bibitem[Bao et~al\mbox{.}(2025)]%
        {surveygen_quality_2025}
\bibfield{author}{\bibinfo{person}{Tong Bao}, \bibinfo{person}{Mir~Tafseer
  Nayeem}, \bibinfo{person}{Davood Rafiei}, {and} \bibinfo{person}{Chengzhi
  Zhang}.} \bibinfo{year}{2025}\natexlab{}.
\newblock \bibinfo{title}{SurveyGen: Quality-Aware Scientific Survey Generation
  with Large Language Models}.
\newblock
\showeprint[arxiv]{2508.17647}~[cs.CL]
\urldef\tempurl%
\url{https://arxiv.org/abs/2508.17647}
\showURL{%
\tempurl}


\bibitem[Bosselut et~al\mbox{.}(2018)]%
        {bosselut2018awardcoherent}
\bibfield{author}{\bibinfo{person}{Antoine Bosselut}, \bibinfo{person}{Asli
  Celikyilmaz}, \bibinfo{person}{Xiaodong He}, \bibinfo{person}{Jianfeng Gao},
  \bibinfo{person}{Po-Sen Huang}, {and} \bibinfo{person}{Yejin Choi}.}
  \bibinfo{year}{2018}\natexlab{}.
\newblock \bibinfo{title}{Discourse-aware neural rewards for coherent text
  generation}.
\newblock


\bibitem[Chang et~al\mbox{.}(2023)]%
        {chang2023survey}
\bibfield{author}{\bibinfo{person}{Yupeng Chang}, \bibinfo{person}{Xu Wang},
  \bibinfo{person}{Jindong Wang}, \bibinfo{person}{Yuan Wu},
  \bibinfo{person}{Linyi Yang}, \bibinfo{person}{Kaijie Zhu},
  \bibinfo{person}{Hao Chen}, \bibinfo{person}{Xiaoyuan Yi},
  \bibinfo{person}{Cunxiang Wang}, \bibinfo{person}{Yidong Wang},
  {et~al\mbox{.}}} \bibinfo{year}{2023}\natexlab{}.
\newblock \bibinfo{title}{A survey on evaluation of large language models}.
\newblock


\bibitem[Chen et~al\mbox{.}(2023c)]%
        {chen2023unleashingprompts}
\bibfield{author}{\bibinfo{person}{Banghao Chen}, \bibinfo{person}{Zhaofeng
  Zhang}, \bibinfo{person}{Nicolas Langrené}, {and} \bibinfo{person}{Shengxin
  Zhu}.} \bibinfo{year}{2023}\natexlab{c}.
\newblock \bibinfo{title}{Unleashing the Potential of Prompt Engineering in
  Large Language Models: A Comprehensive Review}.
\newblock
\showeprint[arxiv]{2310.14735}~[cs.CL]
\urldef\tempurl%
\url{https://arxiv.org/abs/2310.14735}
\showURL{%
\tempurl}


\bibitem[Chen et~al\mbox{.}(2023b)]%
        {chen2023extending}
\bibfield{author}{\bibinfo{person}{Shouyuan Chen}, \bibinfo{person}{Sherman
  Wong}, \bibinfo{person}{Liangjian Chen}, {and} \bibinfo{person}{Yuandong
  Tian}.} \bibinfo{year}{2023}\natexlab{b}.
\newblock \bibinfo{title}{Extending context window of large language models via
  positional interpolation}.
\newblock


\bibitem[Chen et~al\mbox{.}(2023a)]%
        {chen2023longlora}
\bibfield{author}{\bibinfo{person}{Yukang Chen}, \bibinfo{person}{Shengju
  Qian}, \bibinfo{person}{Haotian Tang}, \bibinfo{person}{Xin Lai},
  \bibinfo{person}{Zhijian Liu}, \bibinfo{person}{Song Han}, {and}
  \bibinfo{person}{Jiaya Jia}.} \bibinfo{year}{2023}\natexlab{a}.
\newblock \bibinfo{title}{LongLoRA: Efficient Fine-tuning of Long-Context Large
  Language Models}.
\newblock


\bibitem[Cho et~al\mbox{.}(2018)]%
        {cho2018traingcohernt}
\bibfield{author}{\bibinfo{person}{Woon~Sang Cho}, \bibinfo{person}{Pengchuan
  Zhang}, \bibinfo{person}{Yizhe Zhang}, \bibinfo{person}{Xiujun Li},
  \bibinfo{person}{Michel Galley}, \bibinfo{person}{Chris Brockett},
  \bibinfo{person}{Mengdi Wang}, {and} \bibinfo{person}{Jianfeng Gao}.}
  \bibinfo{year}{2018}\natexlab{}.
\newblock \bibinfo{title}{Towards coherent and cohesive long-form text
  generation}.
\newblock


\bibitem[Dominguez-Olmedo et~al\mbox{.}(2023)]%
        {auto5}
\bibfield{author}{\bibinfo{person}{Ricardo Dominguez-Olmedo},
  \bibinfo{person}{Moritz Hardt}, {and} \bibinfo{person}{Celestine Mendler-D{\"
  u}nner}.} \bibinfo{year}{2023}\natexlab{}.
\newblock \bibinfo{title}{Questioning the {Survey} {Responses} of {Large}
  {Language} {Models}}.
\newblock
\href{https://doi.org/10.48550/ARXIV.2306.07951}{doi:\nolinkurl{10.48550/ARXIV.2306.07951}}


\bibitem[Fang et~al\mbox{.}(2020)]%
        {fang2020beyondlexical}
\bibfield{author}{\bibinfo{person}{Kuan Fang}, \bibinfo{person}{Long Zhao},
  \bibinfo{person}{Zhan Shen}, \bibinfo{person}{RuiXing Wang},
  \bibinfo{person}{RiKang Zhour}, {and} \bibinfo{person}{LiWen Fan}.}
  \bibinfo{year}{2020}\natexlab{}.
\newblock \bibinfo{title}{Beyond Lexical: A Semantic Retrieval Framework for
  Textual Search Engine}.
\newblock
\urldef\tempurl%
\url{https://arxiv.org/abs/2008.03917}
\showURL{%
\tempurl}


\bibitem[Gao et~al\mbox{.}(2024)]%
        {sc8}
\bibfield{author}{\bibinfo{person}{Fan Gao}, \bibinfo{person}{Hang Jiang},
  \bibinfo{person}{Rui Yang}, \bibinfo{person}{Qingcheng Zeng},
  \bibinfo{person}{Jinghui Lu}, \bibinfo{person}{Moritz Blum},
  \bibinfo{person}{Tianwei She}, \bibinfo{person}{Yuang Jiang}, {and}
  \bibinfo{person}{Irene Li}.} \bibinfo{year}{2024}\natexlab{}.
\newblock \bibinfo{title}{Evaluating {Large} {Language} {Models} on
  {Wikipedia}-{Style} {Survey} {Generation}}.
\newblock \bibinfo{numpages}{5405--5418}~pages.
\newblock
\href{https://doi.org/10.18653/v1/2024.findings-acl.321}{doi:\nolinkurl{10.18653/v1/2024.findings-acl.321}}


\bibitem[Gao et~al\mbox{.}(2023)]%
        {gao2023retrieval}
\bibfield{author}{\bibinfo{person}{Yunfan Gao}, \bibinfo{person}{Yun Xiong},
  \bibinfo{person}{Xinyu Gao}, \bibinfo{person}{Kangxiang Jia},
  \bibinfo{person}{Jinliu Pan}, \bibinfo{person}{Yuxi Bi}, \bibinfo{person}{Yi
  Dai}, \bibinfo{person}{Jiawei Sun}, {and} \bibinfo{person}{Haofen Wang}.}
  \bibinfo{year}{2023}\natexlab{}.
\newblock \bibinfo{title}{Retrieval-augmented generation for large language
  models: A survey}.
\newblock


\bibitem[Goodfellow et~al\mbox{.}(2016)]%
        {goodfellow2016deep}
\bibfield{author}{\bibinfo{person}{Ian Goodfellow}, \bibinfo{person}{Yoshua
  Bengio}, {and} \bibinfo{person}{Aaron Courville}.}
  \bibinfo{year}{2016}\natexlab{}.
\newblock \bibinfo{title}{Deep learning}.
\newblock


\bibitem[Ji et~al\mbox{.}(2023)]%
        {ji2023survey}
\bibfield{author}{\bibinfo{person}{Ziwei Ji}, \bibinfo{person}{Nayeon Lee},
  \bibinfo{person}{Rita Frieske}, \bibinfo{person}{Tiezheng Yu},
  \bibinfo{person}{Dan Su}, \bibinfo{person}{Yan Xu}, \bibinfo{person}{Etsuko
  Ishii}, \bibinfo{person}{Ye~Jin Bang}, \bibinfo{person}{Andrea Madotto},
  {and} \bibinfo{person}{Pascale Fung}.} \bibinfo{year}{2023}\natexlab{}.
\newblock \bibinfo{title}{Survey of hallucination in natural language
  generation}.
\newblock \bibinfo{numpages}{38}~pages.
\newblock


\bibitem[Jiang et~al\mbox{.}(2019)]%
        {auto6}
\bibfield{author}{\bibinfo{person}{Xiao-Jian Jiang}, \bibinfo{person}{Xian-Ling
  Mao}, \bibinfo{person}{Bo-Si Feng}, \bibinfo{person}{Xiaochi Wei},
  \bibinfo{person}{Bin-Bin Bian}, {and} \bibinfo{person}{Heyan Huang}.}
  \bibinfo{year}{2019}\natexlab{}.
\newblock \bibinfo{title}{HSDS: An {Abstractive} {Model} for {Automatic}
  {Survey} {Generation}}.
\newblock \bibinfo{numpages}{70--86}~pages.
\newblock
\showISBNx{9783030185756}
\showISSN{0302-9743}
\href{https://doi.org/10.1007/978-3-030-18576-3_5}{doi:\nolinkurl{10.1007/978-3-030-18576-3_5}}


\bibitem[Jiang et~al\mbox{.}(2023)]%
        {jiang2023active}
\bibfield{author}{\bibinfo{person}{Zhengbao Jiang}, \bibinfo{person}{Frank~F
  Xu}, \bibinfo{person}{Luyu Gao}, \bibinfo{person}{Zhiqing Sun},
  \bibinfo{person}{Qian Liu}, \bibinfo{person}{Jane Dwivedi-Yu},
  \bibinfo{person}{Yiming Yang}, \bibinfo{person}{Jamie Callan}, {and}
  \bibinfo{person}{Graham Neubig}.} \bibinfo{year}{2023}\natexlab{}.
\newblock \bibinfo{title}{Active Retrieval Augmented Generation}.
\newblock \bibinfo{numpages}{7969--7992}~pages.
\newblock


\bibitem[Joos et~al\mbox{.}(2024)]%
        {sc1}
\bibfield{author}{\bibinfo{person}{Lucas Joos}, \bibinfo{person}{Daniel~A.
  Keim}, {and} \bibinfo{person}{Maximilian~T. Fischer}.}
  \bibinfo{year}{2024}\natexlab{}.
\newblock \bibinfo{title}{Cutting {Through} the {Clutter}: The {Potential} of
  {LLMs} for {Efficient} {Filtration} in {Systematic} {Literature} {Reviews}}.
\newblock
\href{https://doi.org/10.48550/ARXIV.2407.10652}{doi:\nolinkurl{10.48550/ARXIV.2407.10652}}


\bibitem[Kaddour et~al\mbox{.}(2023)]%
        {kaddour2023challenges}
\bibfield{author}{\bibinfo{person}{Jean Kaddour}, \bibinfo{person}{Joshua
  Harris}, \bibinfo{person}{Maximilian Mozes}, \bibinfo{person}{Herbie
  Bradley}, \bibinfo{person}{Roberta Raileanu}, {and} \bibinfo{person}{Robert
  McHardy}.} \bibinfo{year}{2023}\natexlab{}.
\newblock \bibinfo{title}{Challenges and applications of large language
  models}.
\newblock


\bibitem[Kang and Xiong(2024)]%
        {sc4}
\bibfield{author}{\bibinfo{person}{Hao Kang} {and} \bibinfo{person}{Chenyan
  Xiong}.} \bibinfo{year}{2024}\natexlab{}.
\newblock \bibinfo{title}{ResearchArena: Benchmarking {Large} {Language}
  {Models}' {Ability} to {Collect} and {Organize} {Information} as {Research}
  {Agents}}.
\newblock
\href{https://doi.org/10.48550/ARXIV.2406.10291}{doi:\nolinkurl{10.48550/ARXIV.2406.10291}}


\bibitem[Ke and Ng(2025)]%
        {sc2}
\bibfield{author}{\bibinfo{person}{Ping~Fan Ke} {and} \bibinfo{person}{Ka~Chung
  Ng}.} \bibinfo{year}{2025}\natexlab{}.
\newblock \bibinfo{title}{Human-{AI} {Synergy} in {Survey} {Development}:
  Implications from {Large} {Language} {Models} in {Business} and {Research}}.
\newblock \bibinfo{numpages}{39}~pages.
\newblock
\showISSN{2158-656X}
\href{https://doi.org/10.1145/3700597}{doi:\nolinkurl{10.1145/3700597}}


\bibitem[Khan et~al\mbox{.}(2022)]%
        {khan2022transformers}
\bibfield{author}{\bibinfo{person}{Salman Khan}, \bibinfo{person}{Muzammal
  Naseer}, \bibinfo{person}{Munawar Hayat}, \bibinfo{person}{Syed~Waqas Zamir},
  \bibinfo{person}{Fahad~Shahbaz Khan}, {and} \bibinfo{person}{Mubarak Shah}.}
  \bibinfo{year}{2022}\natexlab{}.
\newblock \bibinfo{title}{Transformers in vision: A survey}.
\newblock \bibinfo{numpages}{41}~pages.
\newblock


\bibitem[Khuong and Rachmat(2023)]%
        {auto2}
\bibfield{author}{\bibinfo{person}{Thanh Gia~Hieu Khuong} {and}
  \bibinfo{person}{Benedictus~Kent Rachmat}.} \bibinfo{year}{2023}\natexlab{}.
\newblock \bibinfo{title}{Auto-survey {Challenge}}.
\newblock
\href{https://doi.org/10.48550/ARXIV.2310.04480}{doi:\nolinkurl{10.48550/ARXIV.2310.04480}}


\bibitem[Kirillov et~al\mbox{.}(2023)]%
        {kirillov2023segment}
\bibfield{author}{\bibinfo{person}{Alexander Kirillov}, \bibinfo{person}{Eric
  Mintun}, \bibinfo{person}{Nikhila Ravi}, \bibinfo{person}{Hanzi Mao},
  \bibinfo{person}{Chloe Rolland}, \bibinfo{person}{Laura Gustafson},
  \bibinfo{person}{Tete Xiao}, \bibinfo{person}{Spencer Whitehead},
  \bibinfo{person}{Alexander~C Berg}, \bibinfo{person}{Wan-Yen Lo},
  {et~al\mbox{.}}} \bibinfo{year}{2023}\natexlab{}.
\newblock \bibinfo{title}{Segment anything}.
\newblock \bibinfo{numpages}{4015--4026}~pages.
\newblock


\bibitem[Kitadai et~al\mbox{.}(2024)]%
        {sc3}
\bibfield{author}{\bibinfo{person}{Ayato Kitadai}, \bibinfo{person}{Kazuhito
  Ogawa}, {and} \bibinfo{person}{Nariaki Nishino}.}
  \bibinfo{year}{2024}\natexlab{}.
\newblock \bibinfo{title}{Examining the {Feasibility} of {Large} {Language}
  {Models} as {Survey} {Respondents}}.
\newblock \bibinfo{numpages}{3858--3864}~pages.
\newblock
\href{https://doi.org/10.1109/bigdata62323.2024.10825497}{doi:\nolinkurl{10.1109/bigdata62323.2024.10825497}}


\bibitem[Lai et~al\mbox{.}(2024a)]%
        {lai2024stepbystep}
\bibfield{author}{\bibinfo{person}{Ting-Han Lai}, \bibinfo{person}{Chia-Wei
  Liu}, {and} \bibinfo{person}{Yun-Nung Chen}.}
  \bibinfo{year}{2024}\natexlab{a}.
\newblock \bibinfo{title}{Step-by-Step Survey Generation: Benchmarking LLMs on
  Structured Scientific Writing Tasks}.
\newblock
\urldef\tempurl%
\url{https://arxiv.org/abs/2407.15906}
\showURL{%
\tempurl}


\bibitem[Lai et~al\mbox{.}(2024b)]%
        {instruct_llm_survey_2024}
\bibfield{author}{\bibinfo{person}{Yuxuan Lai}, \bibinfo{person}{Yupeng Wu},
  \bibinfo{person}{Yidan Wang}, \bibinfo{person}{Wenpeng Hu}, {and}
  \bibinfo{person}{Chen Zheng}.} \bibinfo{year}{2024}\natexlab{b}.
\newblock \bibinfo{title}{Instruct Large Language Models to Generate Scientific
  Literature Survey Step by Step}.
\newblock
\showeprint[arxiv]{2408.07884}~[cs.CL]
\urldef\tempurl%
\url{https://arxiv.org/abs/2408.07884}
\showURL{%
\tempurl}


\bibitem[LeCun et~al\mbox{.}(2015)]%
        {lecun2015deep}
\bibfield{author}{\bibinfo{person}{Yann LeCun}, \bibinfo{person}{Yoshua
  Bengio}, {and} \bibinfo{person}{Geoffrey Hinton}.}
  \bibinfo{year}{2015}\natexlab{}.
\newblock \bibinfo{title}{Deep learning}.
\newblock \bibinfo{numpages}{436--444}~pages.
\newblock


\bibitem[Lee et~al\mbox{.}(2025)]%
        {outline_guided_generation_2025}
\bibfield{author}{\bibinfo{person}{Yukyung Lee}, \bibinfo{person}{Soonwon Ka},
  \bibinfo{person}{Bokyung Son}, \bibinfo{person}{Pilsung Kang}, {and}
  \bibinfo{person}{Jaewook Kang}.} \bibinfo{year}{2025}\natexlab{}.
\newblock \bibinfo{title}{Navigating the Path of Writing: Outline-guided Text
  Generation with Large Language Models}.
\newblock
\showeprint[arxiv]{2404.13919}~[cs.CL]
\urldef\tempurl%
\url{https://arxiv.org/abs/2404.13919}
\showURL{%
\tempurl}


\bibitem[Lewis et~al\mbox{.}(2020)]%
        {lewis2020retrieval}
\bibfield{author}{\bibinfo{person}{Patrick Lewis}, \bibinfo{person}{Ethan
  Perez}, \bibinfo{person}{Aleksandra Piktus}, \bibinfo{person}{Fabio Petroni},
  \bibinfo{person}{Vladimir Karpukhin}, \bibinfo{person}{Naman Goyal},
  \bibinfo{person}{Heinrich K{\"u}ttler}, \bibinfo{person}{Mike Lewis},
  \bibinfo{person}{Wen-tau Yih}, \bibinfo{person}{Tim Rockt{\"a}schel},
  {et~al\mbox{.}}} \bibinfo{year}{2020}\natexlab{}.
\newblock \bibinfo{title}{Retrieval-augmented generation for
  knowledge-intensive nlp tasks}.
\newblock \bibinfo{numpages}{9459--9474}~pages.
\newblock


\bibitem[Li et~al\mbox{.}(2023)]%
        {li2023long}
\bibfield{author}{\bibinfo{person}{Dacheng Li}, \bibinfo{person}{Rulin Shao},
  \bibinfo{person}{Anze Xie}, \bibinfo{person}{Ying Sheng},
  \bibinfo{person}{Lianmin Zheng}, \bibinfo{person}{Joseph Gonzalez},
  \bibinfo{person}{Ion Stoica}, \bibinfo{person}{Xuezhe Ma}, {and}
  \bibinfo{person}{Hao Zhang}.} \bibinfo{year}{2023}\natexlab{}.
\newblock \bibinfo{title}{How Long Can Context Length of Open-Source LLMs truly
  Promise?}
\newblock


\bibitem[Li et~al\mbox{.}(2024)]%
        {li2024long}
\bibfield{author}{\bibinfo{person}{Tianle Li}, \bibinfo{person}{Ge Zhang},
  \bibinfo{person}{Quy~Duc Do}, \bibinfo{person}{Xiang Yue}, {and}
  \bibinfo{person}{Wenhu Chen}.} \bibinfo{year}{2024}\natexlab{}.
\newblock \bibinfo{title}{Long-context LLMs Struggle with Long In-context
  Learning}.
\newblock


\bibitem[Li(2023)]%
        {li2023practicalzeroshot}
\bibfield{author}{\bibinfo{person}{Yinheng Li}.}
  \bibinfo{year}{2023}\natexlab{}.
\newblock \bibinfo{title}{A Practical Survey on Zero-shot Prompt Design for
  In-context Learning}.
\newblock
\showeprint[arxiv]{2309.13205}~[cs.CL]
\urldef\tempurl%
\url{https://arxiv.org/abs/2309.13205}
\showURL{%
\tempurl}


\bibitem[Liu et~al\mbox{.}(2024b)]%
        {liu2024lost}
\bibfield{author}{\bibinfo{person}{Nelson~F Liu}, \bibinfo{person}{Kevin Lin},
  \bibinfo{person}{John Hewitt}, \bibinfo{person}{Ashwin Paranjape},
  \bibinfo{person}{Michele Bevilacqua}, \bibinfo{person}{Fabio Petroni}, {and}
  \bibinfo{person}{Percy Liang}.} \bibinfo{year}{2024}\natexlab{b}.
\newblock \bibinfo{title}{Lost in the middle: How language models use long
  contexts}.
\newblock \bibinfo{numpages}{157--173}~pages.
\newblock


\bibitem[Liu et~al\mbox{.}(2024a)]%
        {liu2024relevancefiltering}
\bibfield{author}{\bibinfo{person}{Y. Liu} {et~al\mbox{.}}}
  \bibinfo{year}{2024}\natexlab{a}.
\newblock \bibinfo{title}{Relevance Filtering for Embedding‐based Retrieval}.
\newblock
\urldef\tempurl%
\url{https://arxiv.org/abs/2408.04887}
\showURL{%
\tempurl}


\bibitem[Maiorino et~al\mbox{.}(2023)]%
        {auto3}
\bibfield{author}{\bibinfo{person}{Antonio Maiorino}, \bibinfo{person}{Zoe
  Padgett}, \bibinfo{person}{Chun Wang}, \bibinfo{person}{Misha Yakubovskiy},
  {and} \bibinfo{person}{Peng Jiang}.} \bibinfo{year}{2023}\natexlab{}.
\newblock \bibinfo{title}{Application and {Evaluation} of {Large} {Language}
  {Models} for the {Generation} of {Survey} {Questions}}.
\newblock \bibinfo{numpages}{5244--5245}~pages.
\newblock
\href{https://doi.org/10.1145/3583780.3615506}{doi:\nolinkurl{10.1145/3583780.3615506}}


\bibitem[Monir et~al\mbox{.}(2024)]%
        {seyedmonir2024vectorsearch}
\bibfield{author}{\bibinfo{person}{Solmaz~Seyed Monir}, \bibinfo{person}{Irene
  Lau}, \bibinfo{person}{Shubing Yang}, {and} \bibinfo{person}{Dongfang Zhao}.}
  \bibinfo{year}{2024}\natexlab{}.
\newblock \bibinfo{title}{VectorSearch: Enhancing Document Retrieval with
  Semantic Embeddings and Optimized Search}.
\newblock
\urldef\tempurl%
\url{https://arxiv.org/abs/2409.17383}
\showURL{%
\tempurl}


\bibitem[Nussbaum et~al\mbox{.}(2025)]%
        {nussbaum2025theoreticallimitations}
\bibfield{author}{\bibinfo{person}{Zach Nussbaum} {et~al\mbox{.}}}
  \bibinfo{year}{2025}\natexlab{}.
\newblock \bibinfo{title}{On the Theoretical Limitations of Embedding‐Based
  Retrieval}.
\newblock
\urldef\tempurl%
\url{https://arxiv.org/abs/2508.21038}
\showURL{%
\tempurl}


\bibitem[Nussbaum and Duderstadt(2025)]%
        {nussbaum2025trainingsparsemixtureexperts}
\bibfield{author}{\bibinfo{person}{Zach Nussbaum} {and}
  \bibinfo{person}{Brandon Duderstadt}.} \bibinfo{year}{2025}\natexlab{}.
\newblock \bibinfo{title}{Training Sparse Mixture of Experts Text Embedding
  Models}.
\newblock
\showeprint[arxiv]{2502.07972}~[cs.CL]
\urldef\tempurl%
\url{https://arxiv.org/abs/2502.07972}
\showURL{%
\tempurl}


\bibitem[Nussbaum et~al\mbox{.}(2024)]%
        {nussbaum2024nomic}
\bibfield{author}{\bibinfo{person}{Zach Nussbaum}, \bibinfo{person}{John~X.
  Morris}, \bibinfo{person}{Brandon Duderstadt}, {and} \bibinfo{person}{Andriy
  Mulyar}.} \bibinfo{year}{2024}\natexlab{}.
\newblock \bibinfo{title}{Nomic Embed: Training a Reproducible Long Context
  Text Embedder}.
\newblock
\showeprint[arxiv]{2402.01613}~[cs.CL]


\bibitem[Pouyanfar et~al\mbox{.}(2018)]%
        {pouyanfar2018survey}
\bibfield{author}{\bibinfo{person}{Samira Pouyanfar}, \bibinfo{person}{Saad
  Sadiq}, \bibinfo{person}{Yilin Yan}, \bibinfo{person}{Haiman Tian},
  \bibinfo{person}{Yudong Tao}, \bibinfo{person}{Maria~Presa Reyes},
  \bibinfo{person}{Mei-Ling Shyu}, \bibinfo{person}{Shu-Ching Chen}, {and}
  \bibinfo{person}{Sundaraja~S Iyengar}.} \bibinfo{year}{2018}\natexlab{}.
\newblock \bibinfo{title}{A survey on deep learning: Algorithms, techniques,
  and applications}.
\newblock \bibinfo{numpages}{36}~pages.
\newblock


\bibitem[Rewina et~al\mbox{.}(2025)]%
        {sc5}
\bibfield{author}{\bibinfo{person}{Bedemariam Rewina}, \bibinfo{person}{Perez
  Natalie}, \bibinfo{person}{Bhaduri S.}, \bibinfo{person}{Kapoor Satya},
  \bibinfo{person}{Gil Alex}, \bibinfo{person}{Conjar Elizabeth},
  \bibinfo{person}{Itoku Ikkei}, \bibinfo{person}{Theil David},
  \bibinfo{person}{Chadha Aman}, {and} \bibinfo{person}{Nayyar Naumaan}.}
  \bibinfo{year}{2025}\natexlab{}.
\newblock \bibinfo{title}{Potential and {Perils} of {Large} {Language} {Models}
  as {Judges} of {Unstructured} {Textual} {Data}}.
\newblock


\bibitem[Sahoo et~al\mbox{.}(2024)]%
        {sahoo2024systematicsurvey}
\bibfield{author}{\bibinfo{person}{Pranab Sahoo}, \bibinfo{person}{Ayush~Kumar
  Singh}, \bibinfo{person}{Sriparna Saha}, \bibinfo{person}{Vinija Jain},
  \bibinfo{person}{Samrat Mondal}, {and} \bibinfo{person}{Aman Chadha}.}
  \bibinfo{year}{2024}\natexlab{}.
\newblock \bibinfo{title}{A Systematic Survey of Prompt Engineering in Large
  Language Models: Techniques and Applications}.
\newblock
\showeprint[arxiv]{2402.07927}~[cs.CL]
\urldef\tempurl%
\url{https://arxiv.org/abs/2402.07927}
\showURL{%
\tempurl}


\bibitem[Shan et~al\mbox{.}(2018)]%
        {shan2018recurrentbinary}
\bibfield{author}{\bibinfo{person}{Ying Shan}, \bibinfo{person}{Jian Jiao},
  \bibinfo{person}{Jie Zhu}, {and} \bibinfo{person}{J.~C. Mao}.}
  \bibinfo{year}{2018}\natexlab{}.
\newblock \bibinfo{title}{Recurrent Binary Embedding for GPU‐Enabled
  Exhaustive Retrieval from Billion‐Scale Semantic Vectors}.
\newblock
\urldef\tempurl%
\url{https://arxiv.org/abs/1802.06466}
\showURL{%
\tempurl}


\bibitem[Shao et~al\mbox{.}(2024)]%
        {shao2024assisting}
\bibfield{author}{\bibinfo{person}{Yijia Shao}, \bibinfo{person}{Yucheng
  Jiang}, \bibinfo{person}{Theodore~A. Kanell}, \bibinfo{person}{Peter Xu},
  \bibinfo{person}{Omar Khattab}, {and} \bibinfo{person}{Monica~S. Lam}.}
  \bibinfo{year}{2024}\natexlab{}.
\newblock \bibinfo{title}{{Assisting in Writing Wikipedia-like miscs From
  Scratch with Large Language Models}}.
\newblock


\bibitem[Shi et~al\mbox{.}(2023)]%
        {shi2023large}
\bibfield{author}{\bibinfo{person}{Freda Shi}, \bibinfo{person}{Xinyun Chen},
  \bibinfo{person}{Kanishka Misra}, \bibinfo{person}{Nathan Scales},
  \bibinfo{person}{David Dohan}, \bibinfo{person}{Ed~H Chi},
  \bibinfo{person}{Nathanael Sch{\"a}rli}, {and} \bibinfo{person}{Denny Zhou}.}
  \bibinfo{year}{2023}\natexlab{}.
\newblock \bibinfo{title}{Large language models can be easily distracted by
  irrelevant context}.
\newblock \bibinfo{numpages}{31210--31227}~pages.
\newblock


\bibitem[Tan et~al\mbox{.}(2024)]%
        {auto4}
\bibfield{author}{\bibinfo{person}{Zhen Tan}, \bibinfo{person}{Dawei Li},
  \bibinfo{person}{Song Wang}, \bibinfo{person}{Alimohammad Beigi},
  \bibinfo{person}{Bohan Jiang}, \bibinfo{person}{Amrita Bhattacharjee},
  \bibinfo{person}{Mansooreh Karami}, \bibinfo{person}{Jundong Li},
  \bibinfo{person}{Lu Cheng}, {and} \bibinfo{person}{Huan Liu}.}
  \bibinfo{year}{2024}\natexlab{}.
\newblock \bibinfo{title}{Large {Language} {Models} for {Data} {Annotation} and
  {Synthesis}: A {Survey}}.
\newblock
\href{https://doi.org/10.48550/ARXIV.2402.13446}{doi:\nolinkurl{10.48550/ARXIV.2402.13446}}


\bibitem[Touvron et~al\mbox{.}(2023)]%
        {touvron2023llama}
\bibfield{author}{\bibinfo{person}{Hugo Touvron}, \bibinfo{person}{Thibaut
  Lavril}, \bibinfo{person}{Gautier Izacard}, \bibinfo{person}{Xavier
  Martinet}, \bibinfo{person}{Marie-Anne Lachaux},
  \bibinfo{person}{Timoth{\'e}e Lacroix}, \bibinfo{person}{Baptiste
  Rozi{\`e}re}, \bibinfo{person}{Naman Goyal}, \bibinfo{person}{Eric Hambro},
  \bibinfo{person}{Faisal Azhar}, {et~al\mbox{.}}}
  \bibinfo{year}{2023}\natexlab{}.
\newblock \bibinfo{title}{Llama: Open and efficient foundation language
  models}.
\newblock


\bibitem[Wan et~al\mbox{.}(2023)]%
        {auto8}
\bibfield{author}{\bibinfo{person}{Zhongwei Wan}, \bibinfo{person}{Xin Wang},
  \bibinfo{person}{Che Liu}, \bibinfo{person}{Samiul Alam}, \bibinfo{person}{Yu
  Zheng}, \bibinfo{person}{Jiachen Liu}, \bibinfo{person}{Zhongnan Qu},
  \bibinfo{person}{Shen Yan}, \bibinfo{person}{Yi Zhu}, \bibinfo{person}{Quanlu
  Zhang}, \bibinfo{person}{Mosharaf Chowdhury}, {and} \bibinfo{person}{Mi
  Zhang}.} \bibinfo{year}{2023}\natexlab{}.
\newblock \bibinfo{title}{Efficient {Large} {Language} {Models}: A {Survey}}.
\newblock
\href{https://doi.org/10.48550/ARXIV.2312.03863}{doi:\nolinkurl{10.48550/ARXIV.2312.03863}}


\bibitem[Wang et~al\mbox{.}(2023)]%
        {wang2023surveyfact}
\bibfield{author}{\bibinfo{person}{Cunxiang Wang}, \bibinfo{person}{Xiaoze
  Liu}, \bibinfo{person}{Yuanhao Yue}, \bibinfo{person}{Xiangru Tang},
  \bibinfo{person}{Tianhang Zhang}, \bibinfo{person}{Cheng Jiayang},
  \bibinfo{person}{Yunzhi Yao}, \bibinfo{person}{Wenyang Gao},
  \bibinfo{person}{Xuming Hu}, \bibinfo{person}{Zehan Qi}, {et~al\mbox{.}}}
  \bibinfo{year}{2023}\natexlab{}.
\newblock \bibinfo{title}{Survey on factuality in large language models:
  Knowledge, retrieval and domain-specificity}.
\newblock


\bibitem[Wang et~al\mbox{.}(2019)]%
        {wang2019paperrobot}
\bibfield{author}{\bibinfo{person}{Qingyun Wang}, \bibinfo{person}{Lifu Huang},
  \bibinfo{person}{Zhiying Jiang}, \bibinfo{person}{Kevin Knight},
  \bibinfo{person}{Heng Ji}, \bibinfo{person}{Mohit Bansal}, {and}
  \bibinfo{person}{Yi Luan}.} \bibinfo{year}{2019}\natexlab{}.
\newblock \bibinfo{title}{PaperRobot: Incremental draft generation of
  scientific ideas}.
\newblock


\bibitem[Wang et~al\mbox{.}(2024a)]%
        {wang2024augmenting}
\bibfield{author}{\bibinfo{person}{Weizhi Wang}, \bibinfo{person}{Li Dong},
  \bibinfo{person}{Hao Cheng}, \bibinfo{person}{Xiaodong Liu},
  \bibinfo{person}{Xifeng Yan}, \bibinfo{person}{Jianfeng Gao}, {and}
  \bibinfo{person}{Furu Wei}.} \bibinfo{year}{2024}\natexlab{a}.
\newblock \bibinfo{title}{Augmenting language models with long-term memory}.
\newblock


\bibitem[Wang et~al\mbox{.}(2024b)]%
        {auto1}
\bibfield{author}{\bibinfo{person}{Yidong Wang}, \bibinfo{person}{Qi Guo},
  \bibinfo{person}{Wenjin Yao}, \bibinfo{person}{Hongbo Zhang},
  \bibinfo{person}{Xin Zhang}, \bibinfo{person}{Zhen Wu},
  \bibinfo{person}{Meishan Zhang}, \bibinfo{person}{Xinyu Dai},
  \bibinfo{person}{Min Zhang}, \bibinfo{person}{Qingsong Wen},
  \bibinfo{person}{Wei Ye}, \bibinfo{person}{Shikun Zhang}, {and}
  \bibinfo{person}{Yue Zhang}.} \bibinfo{year}{2024}\natexlab{b}.
\newblock \bibinfo{title}{AutoSurvey: Large {Language} {Models} {Can}
  {Automatically} {Write} {Surveys}}.
\newblock
\href{https://doi.org/10.48550/ARXIV.2406.10252}{doi:\nolinkurl{10.48550/ARXIV.2406.10252}}


\bibitem[Wang et~al\mbox{.}(2024c)]%
        {wang2024autosurvey}
\bibfield{author}{\bibinfo{person}{Yue Wang}, \bibinfo{person}{Yuhao Wang},
  \bibinfo{person}{Xinyu Hu}, \bibinfo{person}{Jianbo Yuan},
  \bibinfo{person}{Xipeng Qiu}, {and} \bibinfo{person}{Xuanjing Huang}.}
  \bibinfo{year}{2024}\natexlab{c}.
\newblock \bibinfo{title}{AutoSurvey: Large Language Model-Driven Automated
  Survey Generation}.
\newblock
\urldef\tempurl%
\url{https://arxiv.org/abs/2403.06278}
\showURL{%
\tempurl}


\bibitem[Wang et~al\mbox{.}(2024d)]%
        {pandalm2024}
\bibfield{author}{\bibinfo{person}{Yidong Wang}, \bibinfo{person}{Zhuohao Yu},
  \bibinfo{person}{Zhengran Zeng}, \bibinfo{person}{Linyi Yang},
  \bibinfo{person}{Cunxiang Wang}, \bibinfo{person}{Hao Chen},
  \bibinfo{person}{Chaoya Jiang}, \bibinfo{person}{Rui Xie},
  \bibinfo{person}{Jindong Wang}, \bibinfo{person}{Xing Xie},
  \bibinfo{person}{Wei Ye}, \bibinfo{person}{Shikun Zhang}, {and}
  \bibinfo{person}{Yue Zhang}.} \bibinfo{year}{2024}\natexlab{d}.
\newblock \bibinfo{title}{PandaLM: An Automatic Evaluation Benchmark for LLM
  Instruction Tuning Optimization}.
\newblock


\bibitem[Wei et~al\mbox{.}(2025)]%
        {wei2025plangenllms}
\bibfield{author}{\bibinfo{person}{Zexuan Wei}, \bibinfo{person}{Xinyuan Xu},
  \bibinfo{person}{Zihao Fu}, \bibinfo{person}{Yichong Xu},
  \bibinfo{person}{Zhiruo Wang}, \bibinfo{person}{Ruiqi Zhong}, {and}
  \bibinfo{person}{Chenguang Zhu}.} \bibinfo{year}{2025}\natexlab{}.
\newblock \bibinfo{title}{PlanGen: Structuring Long-Form Generation via
  Hierarchical Planning with Large Language Models}.
\newblock
\urldef\tempurl%
\url{https://arxiv.org/abs/2501.04512}
\showURL{%
\tempurl}


\bibitem[Wu et~al\mbox{.}(2025b)]%
        {wu2025sitemb}
\bibfield{author}{\bibinfo{person}{Junjie Wu}, \bibinfo{person}{Jiangnan Li},
  \bibinfo{person}{Yuqing Li}, \bibinfo{person}{Lemao Liu},
  \bibinfo{person}{Liyan Xu}, \bibinfo{person}{Jiwei Li},
  \bibinfo{person}{Dit-Yan Yeung}, \bibinfo{person}{Jie Zhou}, {and}
  \bibinfo{person}{Mo Yu}.} \bibinfo{year}{2025}\natexlab{b}.
\newblock \bibinfo{title}{SitEmb-v1.5: Improved Context-Aware Dense Retrieval
  for Semantic Association and Long Story Comprehension}.
\newblock
\urldef\tempurl%
\url{https://arxiv.org/abs/2508.01959}
\showURL{%
\tempurl}


\bibitem[Wu et~al\mbox{.}(2025a)]%
        {automated_review_generation_2025}
\bibfield{author}{\bibinfo{person}{S. Wu} {et~al\mbox{.}}}
  \bibinfo{year}{2025}\natexlab{a}.
\newblock \bibinfo{title}{Automated Review Generation Method Powered by Large
  Language Models}.
\newblock
\href{https://doi.org/10.1093/nsr/nwaf169}{doi:\nolinkurl{10.1093/nsr/nwaf169}}


\bibitem[Wuttke et~al\mbox{.}(2024)]%
        {sc7}
\bibfield{author}{\bibinfo{person}{Alexander Wuttke}, \bibinfo{person}{Matthias
  A\ss{}enmacher}, \bibinfo{person}{Christopher Klamm}, \bibinfo{person}{Max~M.
  Lang}, \bibinfo{person}{Quirin W{\" u}rschinger}, {and}
  \bibinfo{person}{Frauke Kreuter}.} \bibinfo{year}{2024}\natexlab{}.
\newblock \bibinfo{title}{AI {Conversational} {Interviewing}: Transforming
  {Surveys} with {LLMs} as {Adaptive} {Interviewers}}.
\newblock
\href{https://doi.org/10.48550/ARXIV.2410.01824}{doi:\nolinkurl{10.48550/ARXIV.2410.01824}}


\bibitem[Yan et~al\mbox{.}(2025)]%
        {yan2025surveyforge}
\bibfield{author}{\bibinfo{person}{X. Yan} {et~al\mbox{.}}}
  \bibinfo{year}{2025}\natexlab{}.
\newblock \bibinfo{title}{SURVEYFORGE: On the Outline Heuristics, Memory-Guided
  Retrieval and Generation of Survey Papers}.
\newblock
\urldef\tempurl%
\url{https://aclanthology.org/2025.acl-long.609.pdf}
\showURL{%
\tempurl}


\bibitem[Yu et~al\mbox{.}(2024)]%
        {yu2024kieval}
\bibfield{author}{\bibinfo{person}{Zhuohao Yu}, \bibinfo{person}{Chang Gao},
  \bibinfo{person}{Wenjin Yao}, \bibinfo{person}{Yidong Wang},
  \bibinfo{person}{Wei Ye}, \bibinfo{person}{Jindong Wang},
  \bibinfo{person}{Xing Xie}, \bibinfo{person}{Yue Zhang}, {and}
  \bibinfo{person}{Shikun Zhang}.} \bibinfo{year}{2024}\natexlab{}.
\newblock \bibinfo{title}{KIEval: A Knowledge-grounded Interactive Evaluation
  Framework for Large Language Models}.
\newblock


\bibitem[Zhao et~al\mbox{.}(2023b)]%
        {zhao2023survey}
\bibfield{author}{\bibinfo{person}{Wayne~Xin Zhao}, \bibinfo{person}{Kun Zhou},
  \bibinfo{person}{Junyi Li}, \bibinfo{person}{Tianyi Tang},
  \bibinfo{person}{Xiaolei Wang}, \bibinfo{person}{Yupeng Hou},
  \bibinfo{person}{Yingqian Min}, \bibinfo{person}{Beichen Zhang},
  \bibinfo{person}{Junjie Zhang}, \bibinfo{person}{Zican Dong},
  {et~al\mbox{.}}} \bibinfo{year}{2023}\natexlab{b}.
\newblock \bibinfo{title}{A survey of large language models}.
\newblock


\bibitem[Zhao et~al\mbox{.}(2023a)]%
        {auto7}
\bibfield{author}{\bibinfo{person}{Wayne~Xin Zhao}, \bibinfo{person}{Kun Zhou},
  \bibinfo{person}{Junyi Li}, \bibinfo{person}{Tianyi Tang},
  \bibinfo{person}{Xiaolei Wang}, \bibinfo{person}{Yupeng Hou},
  \bibinfo{person}{Yingqian Min}, \bibinfo{person}{Beichen Zhang},
  \bibinfo{person}{Junjie Zhang}, \bibinfo{person}{Zican Dong},
  \bibinfo{person}{Yifan Du}, \bibinfo{person}{Chen Yang},
  \bibinfo{person}{Yushuo Chen}, \bibinfo{person}{Zhipeng Chen},
  \bibinfo{person}{Jinhao Jiang}, \bibinfo{person}{Ruiyang Ren},
  \bibinfo{person}{Yifan Li}, \bibinfo{person}{Xinyu Tang},
  \bibinfo{person}{Zikang Liu}, \bibinfo{person}{Peiyu Liu},
  \bibinfo{person}{Jian-Yun Nie}, {and} \bibinfo{person}{Ji-Rong Wen}.}
  \bibinfo{year}{2023}\natexlab{a}.
\newblock \bibinfo{title}{A {Survey} of {Large} {Language} {Models}}.
\newblock
\href{https://doi.org/10.48550/ARXIV.2303.18223}{doi:\nolinkurl{10.48550/ARXIV.2303.18223}}


\bibitem[Zheng et~al\mbox{.}(2024)]%
        {zheng2024judging}
\bibfield{author}{\bibinfo{person}{Lianmin Zheng}, \bibinfo{person}{Wei-Lin
  Chiang}, \bibinfo{person}{Ying Sheng}, \bibinfo{person}{Siyuan Zhuang},
  \bibinfo{person}{Zhanghao Wu}, \bibinfo{person}{Yonghao Zhuang},
  \bibinfo{person}{Zi Lin}, \bibinfo{person}{Zhuohan Li},
  \bibinfo{person}{Dacheng Li}, \bibinfo{person}{Eric Xing}, {et~al\mbox{.}}}
  \bibinfo{year}{2024}\natexlab{}.
\newblock \bibinfo{title}{Judging llm-as-a-judge with mt-bench and chatbot
  arena}.
\newblock


\bibitem[Ziming et~al\mbox{.}(2025)]%
        {sc6}
\bibfield{author}{\bibinfo{person}{Luo Ziming}, \bibinfo{person}{Yang Zonglin},
  \bibinfo{person}{Xu Zexin}, \bibinfo{person}{Yang Wei}, {and}
  \bibinfo{person}{Du Xinya}.} \bibinfo{year}{2025}\natexlab{}.
\newblock \bibinfo{title}{LLM4SR: A {Survey} on {Large} {Language} {Models} for
  {Scientific} {Research}}.
\newblock


\end{thebibliography}

\end{document}